\documentclass{article}

\usepackage{microtype}
\usepackage{graphicx}
\usepackage{subfigure}
\usepackage{caption}
\usepackage{booktabs} % for professional tables

\usepackage{hyperref}

\usepackage[accepted]{icml2025}

\usepackage{amsmath}
\usepackage{amssymb}
\usepackage{mathtools}
\usepackage{amsthm}
\usepackage{multirow}
\usepackage{bbding}
\usepackage[capitalize,noabbrev]{cleveref}

\theoremstyle{plain}

\theoremstyle{definition}

\theoremstyle{remark}

\usepackage[textsize=tiny]{todonotes}

\icmltitlerunning{Improving Multimodal Learning Balance and Sufficiency through Data Remixing}

\begin{document}

\twocolumn[
\icmltitle{Improving Multimodal Learning Balance and Sufficiency \\ through Data Remixing}



\begin{icmlauthorlist}
\icmlauthor{Xiaoyu Ma}{sch1,comp}
\icmlauthor{Hao Chen}{sch1,comp}
\icmlauthor{Yongjian Deng}{sch2}

\end{icmlauthorlist}

\icmlaffiliation{sch1}{School of Computer Science and Engineering, Southeast University, Nanjing, China}
\icmlaffiliation{comp}{Key Laboratory of New Generation Artificial Intelligence Technology and Its Interdisciplinary
Applications (Southeast University), Ministry of Education, China}
\icmlaffiliation{sch2}{College of Computer Science, Beijing University of Technology, Beijing, China}

\icmlcorrespondingauthor{Hao Chen}{haochen303@seu.edu.cn}

\icmlkeywords{Multimodal Learning, Representation Learning}

\vskip 0.3in
]



\printAffiliationsAndNotice{}  % leave blank if no need to mention equal contribution

\begin{abstract}
Different modalities hold considerable gaps in optimization trajectories, including speeds and paths, which lead to \textit{modality laziness} and \textit{modality clash} when jointly training multimodal models, resulting in insufficient and imbalanced multimodal learning.
Existing methods focus on enforcing the weak modality by adding modality-specific optimization objectives, aligning their optimization speeds, or decomposing multimodal learning to enhance unimodal learning. These methods fail to achieve both unimodal sufficiency and multimodal balance.
In this paper, we, for the first time, address both concerns by proposing multimodal Data Remixing, including decoupling multimodal data and filtering hard samples for each modality to mitigate modality imbalance; and then batch-level reassembling to align the gradient directions and avoid cross-modal interference, thus enhancing unimodal learning sufficiency. 
Experimental results demonstrate that our method can be seamlessly integrated with existing approaches, improving accuracy by approximately \textbf{6.50\%$\uparrow$} on CREMAD and \textbf{3.41\%$\uparrow$} on Kinetic-Sounds, without training set expansion or additional computational overhead during inference. The source code is available at \href{https://github.com/MatthewMaxy/Remix_ICML2025}{Data Remixing}.
\end{abstract}

\section{Introduction}
Multimodal learning \cite{ngiam2011multimodal} is a rapidly evolving field in artificial intelligence, aimed at enhancing the perception and decision-making capabilities of models by integrating data from diverse modalities, including vision, sound, and text \cite{vision+}. However, existing multimodal learning methods often face challenges in fully integrating rich multimodal knowledge across different modalities.
Due to the inherent differences in data representation and distribution across modalities, their optimization trajectories differ significantly, resulting in imbalanced learning when multiple modalities are jointly trained under a unified objective.
\begin{figure}[t]
\vskip 0.2in
\begin{center}
\centerline{\includegraphics[width=1.08\columnwidth]{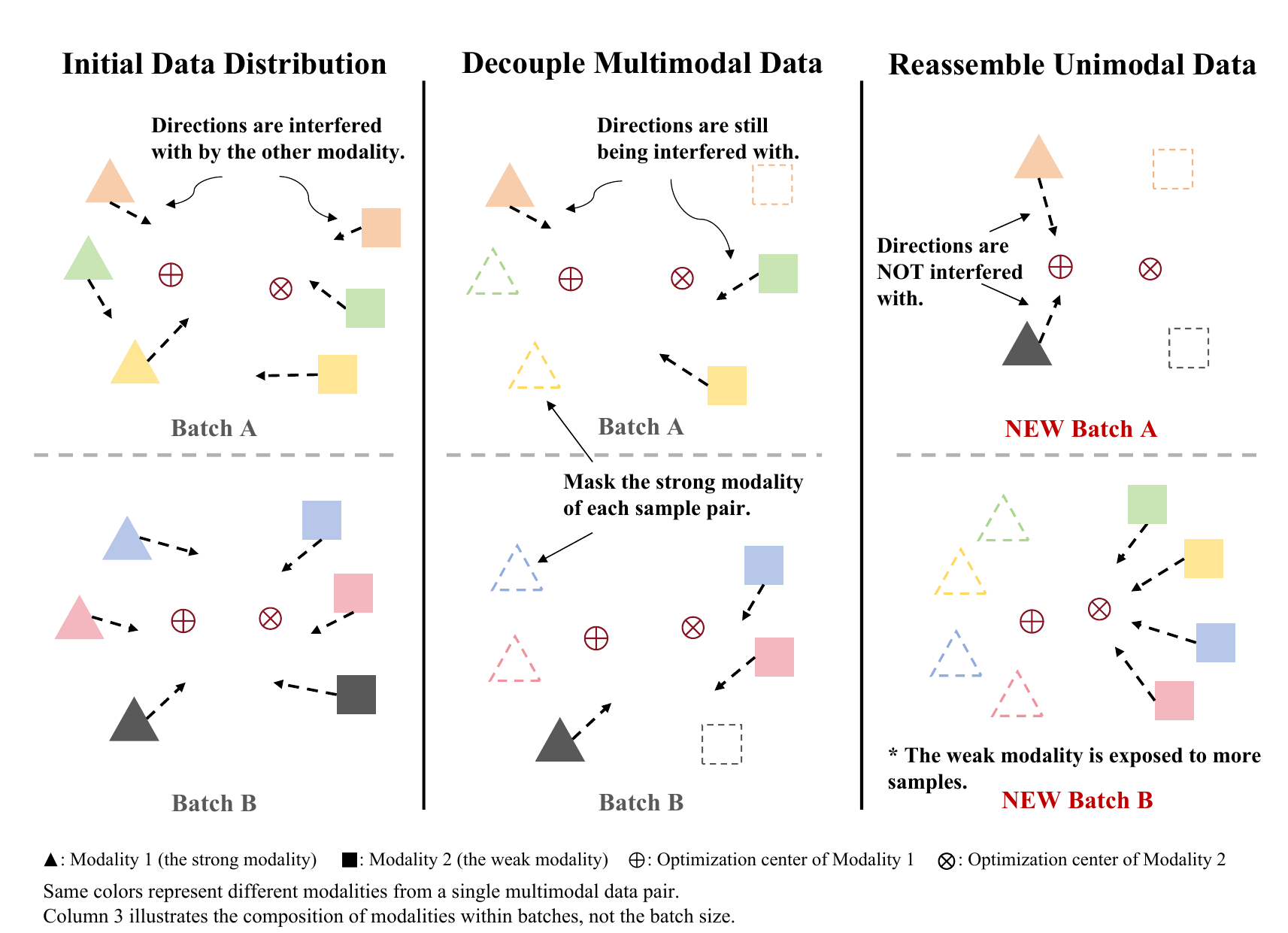}}
\caption{We decouple the multimodal data to assign samples to each modality’s training and then reassemble the inputs to control the consistency of modalities within the batch. By regulating the number of samples, we mitigate modality laziness, and by adjusting the batch composition, we alleviate modality clash.}
\label{fig:motivation}
\end{center}
\vskip -0.4in
\end{figure}
Specifically, multimodal models often prioritize learning the most discriminative features from the strong modality, suppressing the training of the weak modality and causing it to become lazy, known as \textit{modality laziness} \cite{wang2020makes, huang2022modality, du2021improving}.

Several methods have been proposed to address these issues. Some focus on alleviating modality laziness by enforcing the weak modality through modality-specific optimization objectives, such as adjusting task-specific supervision \cite{wang2020makes, xu2023mmcosine, du2023uni}, introducing prototype learning \cite{fan2023pmr}, or using knowledge distillation \cite{du2021improving}.
Other methods aim to align optimization speeds by modifying learning rates \cite{sun2021learning} and gradients \cite{peng2022balanced, li2023boosting, sun2023graph, fu2023multimodal, kontras2024improving, wei2024mmpareto} based on unimodal performance.
In contrast, some approaches attempt to regulate modality inputs to more explicitly enhance the learning of weak modalities, such as by augmenting weak modality samples \cite{greedy, wei2024enhancing} and masking strong ones \cite{zhou2023adaptive}, or by decoupling multimodal learning into unimodal tasks to avoid modality laziness \cite{mla}.
However, most of the existing methods align the optimization speeds of different modalities to alleviate modality laziness, without addressing the fact that even when modality balance is achieved, differing optimization paths can still cause interference during modality optimization.
In multimodal learning, the gradient update directions of different modalities are inconsistent \cite{fan2023pmr}, leading to cross-modal interference during batch gradient descent (Figure \ref{fig:angle}), which can be called \textit{modality clash}. This inconsistency causes the modalities to deviate from their expected optimization paths, as shown in Figure \ref{fig:motivation}, resulting in insufficient learning across all modalities and limiting the effectiveness of multimodal learning.
Moreover, some methods are constrained by specific model architectures \cite{he2022multimodal, lin2023variational, mla} or hinder training efficiency \cite{wei2024enhancing, mla}, which further limits their applicability.

\begin{figure*}[ht]
\vskip 0.2in
\begin{center}
\centerline{\includegraphics[width=\textwidth]{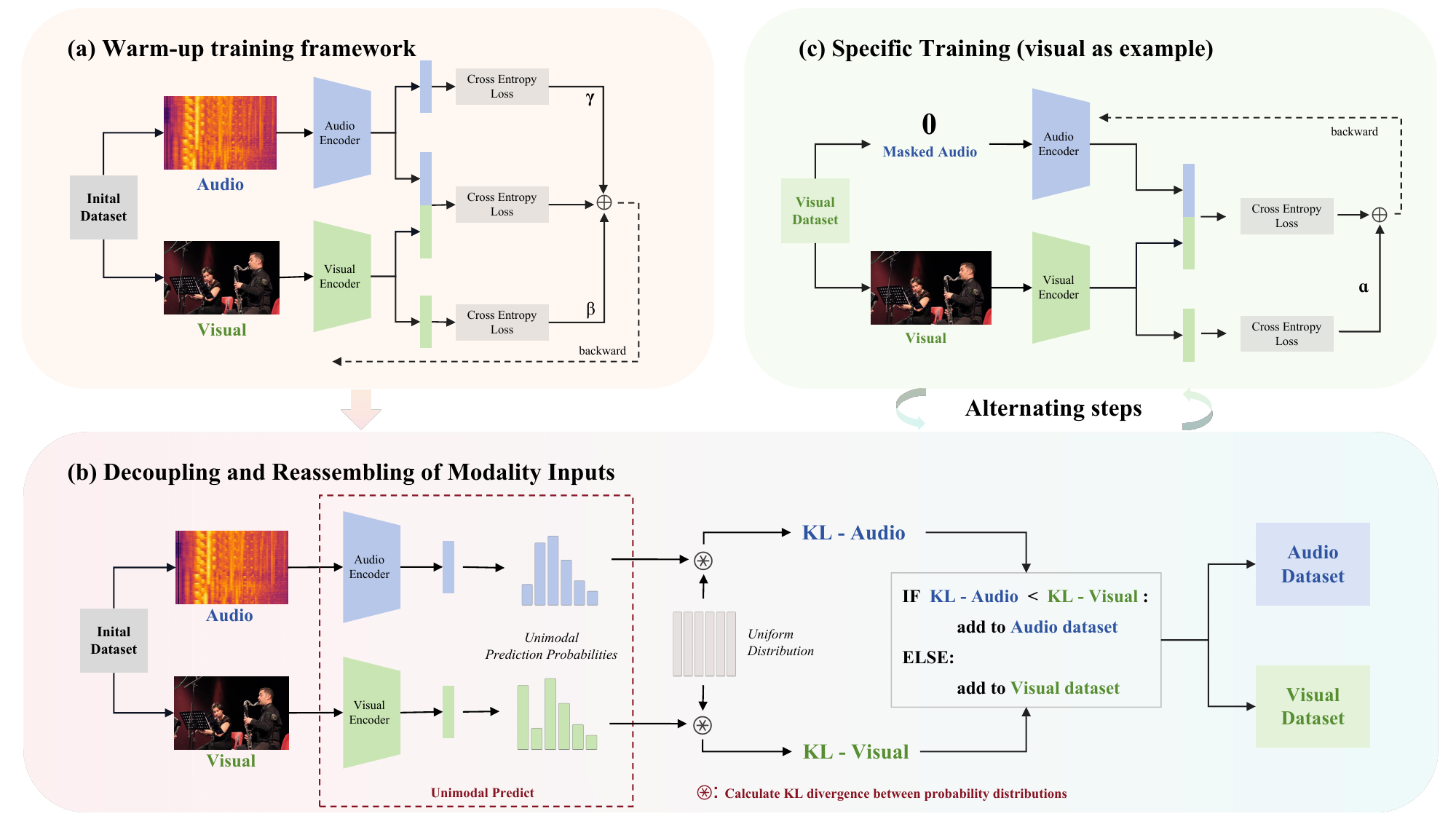}}
\caption{The pipeline of Data Remixing method. In step (a), the complete dataset is used for training to ensure the model develops the basic representational capability. In step (b), the multimodal data is decoupled based on unimodal separability (calculated using KL-divergence), and the original dataset is reassembled into non-overlapping subsets. In step (c), the subsets obtained in step (b) are used to train on the specific modality by masking other modalities to zero.}
\label{fig:pipeline}
\end{center}
\vskip -0.2in
\end{figure*}

Recognizing these limitations, we propose the \textbf{Data Remixing} method, illustrated in Figure \ref{fig:pipeline}. Through dynamic sample allocation and batch-level alignment mechanisms, we simultaneously address modality laziness and modality clash without hindering training efficiency or being constrained by specific model architectures.

Concretely, our Data Remixing method consists of two key steps: sample-level decoupling of multimodal data and batch-level reassembling of unimodal data.
First, based on unimodal separability, we evaluate the representational capability of each modality at the sample level, retaining only the input from the weak modality while masking the others to zero. By decoupling the multimodal data, multimodal models can leverage specific samples to train each modality effectively.
However, even after decoupling, unimodal gradient update directions still lead to interference. We further analyze the optimization process and propose that modality clash originates at the batch level. To address this, we reassemble unimodal data based on the decoupling assignments, ensuring that each batch contains data from only one modality, free from interference from other modalities, as shown in column 3 of Figure \ref{fig:motivation}. Such sample-level decoupling and batch-level reassembling allow each modality to be learned sufficiently and balanced, all without expanding the dataset. 

We test our method on different datasets and achieve excellent results. When combined with conventional fusion methods, the model’s performance shows a notable improvement across multiple datasets. Importantly, our strategy does not expand the dataset or introduce additional overhead during inference. We summarize our key contributions as follows:
\begin{itemize}
    \item We introduce the Data Remixing method, a training strategy that combines decoupling multimodal data and reassembling unimodal data to address modality laziness and clash. Our method does not require dataset expansion and is not constrained by specific model architectures.
    \item We analyze the optimization process and propose that modality laziness and clash originate at the batch level. To the best of our knowledge, we are the first to analyze the challenges of multimodal learning at the batch level and propose a solution.
    \item We perform experiments and demonstrate that (1) Data Remixing achieves excellent results in multimodal learning tasks; (2) Data Remixing can be easily and effectively integrated with other methods, leading to significant improvements of approximately \textbf{6.50\%$\uparrow$} on CREMAD and \textbf{3.41\%$\uparrow$} on Kinetic-Sounds.
\end{itemize}

\section{Related Work}

In this section, we introduce several strategies for multimodal balance learning. We categorize these methods based on the form of data input. The methods described in Section \ref{sec:mu} retain multimodal inputs throughout the learning process, whereas those in Section \ref{sec:uni} operate with unimodal inputs (though not necessarily for the entire duration of training).

\subsection{Balance with Multimodal Joint Input}\label{sec:mu}
\citet{wang2020makes,du2021improving} discover that the optimization speed varies across different modalities. Therefore, when optimizing a multimodal model with a unified objective, the weak modality fails to learn adequately, leading to modality laziness. Most methods maintain multimodal data inputs during training, achieving modality balance by adjusting optimization objectives and speeds.

Some try to alleviate modality laziness by enforcing the weak modality through modality-specific optimization objectives \citet{wang2020makes, du2021improving, xu2023mmcosine, du2023uni, fan2023pmr}. \citet{wang2020makes} propose Gradient-Blending, which calculates the optimal mixing mode of modality losses by determining the overfitting situation for each modality. \citet{du2021improving} attempt to distill knowledge from well-trained unimodal models to enhance the unimodal encoders. \citet{fan2023pmr} introduce the prototype cross-entropy loss for each modality to accelerate the slow-learning modality.
Others try to align optimization speed by changing learning rates \cite{sun2021learning} or modifying gradients  \cite{peng2022balanced, li2023boosting, sun2023graph, fu2023multimodal, kontras2024improving} according to unimodal performance. \citet{sun2021learning} dynamically adjust the learning rates of different modalities based on the unimodal predictive loss. \citet{peng2022balanced} adaptively modulate the gradients of each modality by monitoring the discrepancy in their contribution to the learning objective. These methods promote modality balance and enhance the expressive power of multimodal models. However, they fail to recognize that even when modality laziness is addressed, modality clash persists, meaning that multimodal capability remains limited.

\subsection{Balance with Unimodal Single Input}\label{sec:uni}
Some methods attempt to use alternating or selective unimodal inputs to promote multimodal learning, thereby avoiding interference from other modalities and enhancing the expressive power of multimodal models \cite{greedy, mla, wei2024enhancing, zhou2023adaptive}.
\citet{greedy} measure the relative speed of each modality and train unimodal branches to fully utilize them.
\citet{mla} decompose multimodal learning into alternating unimodal learning with gradient modification to preserve cross-modal interactions. 
\citet{wei2024enhancing} design a sample-level modality contribution evaluation based on Shapley values and perform resampling of the weak modality to enhance specific modality training. 
These methods, which rely on unimodal inputs, seem to avoid modality clash. However, upon closer inspection, we find that modality clash is introduced at the batch level. Like the methods in Section \ref{sec:mu}, they have not analyzed or addressed the root cause of modality clash.

Overall, current work focuses on addressing modality laziness but fails to recognize that modality clash still exists in balanced multimodal learning. Meanwhile, some methods are constrained by specific model architectures \cite{greedy, mla} or hinder training efficiency \cite{mla, wei2024enhancing}, limiting their applicability.
In this paper, we aim to simultaneously mitigate modality laziness and modality clash while ensuring compatibility with various fusion methods and model architectures.

\section{Method}
\subsection{Model formulation}
In this paper, we focus on the multimodal discrimination task, following related works \cite{peng2022balanced, fan2023pmr, wei2024enhancing, mla}. For convenience, we consider two input modalities: $m_a$ and $m_v$.
The training dataset is $\mathcal{D} = \{x_i, y_i\}_{i=1,2,\ldots,N}$. Each $x_i$ consists of multimodal inputs, i.e., $x_i = (x_i^a, x_i^v)$. $y_i \in \{1, 2, \ldots, M\}$, where $M$ is the number of classes. The multimodal model is trained using batch gradient descent, with data $\boldsymbol{x} = (\boldsymbol{x}^a, \boldsymbol{x}^v)$ sampled from a batch $B$.

We use a multimodal model consisting of two unimodal branches for prediction. Each branch has a unimodal encoder, denoted as $\phi^a$ and $\phi^v$, used to extract features from the corresponding modality of $\boldsymbol{x}$. The encoder outputs are represented as $\boldsymbol{z}^a = \phi^a(\theta^a, \boldsymbol{x}^a)$ and $\boldsymbol{z}^v = \phi^v(\theta^v, \boldsymbol{x}^v)$, where $\theta^a$ and $\theta^v$ are the parameters of the encoders. The results of the two unimodal encoders are fused in some way \cite{concat, dicisionfuse} to obtain the multimodal output. We use cross-entropy (CE) loss as the loss function. To achieve better representation ability in the warm-up stage, we add a separate classification head for each modality, as \citet{wang2020makes} do, and modify the loss function as follows:
\begin{equation}
\label{equ:loss}
   \mathcal{L} = \mathcal{L}_{CE}^0 + \sum_{k=1}^K w_k \mathcal{L}_{CE}^{m_k},
\end{equation}

where $\mathcal{L}_{CE}^0$ represents the CE loss for multimodal prediction, while $\mathcal{L}_{CE}^{m_k}$ represents the CE loss for the individual prediction of the $k$-th modality. We provide experiments to demonstrate that the improvements brought by our method are not dependent on the design of this loss function.

\subsection{Data Remixing Method}\label{sec:remix}
\textbf{Overview of Method.}
The complete pipeline of our method is presented in Figure \ref{fig:pipeline}.
In step (a), we first use the complete dataset and multimodal inputs for warm-up training to ensure the model has the basic representational capability. Then, the model is optimized through alternating steps (b) and (c). In step (b), we decouple the multimodal inputs based on the KL divergence of unimodal prediction probabilities and reassemble the data at the batch level according to the remaining modality. In step (c), we perform specific training for each modality using the reassembled dataset.

\textbf{Decouple Multimodal Data.}
In multimodal learning, the model tends to fit the modality that optimizes faster, often converging before the other modality has been sufficiently learned \cite{du2021improving}. Therefore, appropriately reducing the training speed of the strong modality may be an effective strategy for balancing learning \cite{peng2022balanced}. However, modality-level control over the optimization process or objectives overlooks the imbalance at the sample level between modalities \cite{wei2024enhancing}, which limits the effectiveness of the balancing. Based on this premise, we consider promoting balance by assigning specific samples (in terms of difficulty and quantity) to different modalities, achieving a more precise and convenient modality balance.

We first define a sample-level unimodal capability evaluation method to achieve either-or data allocation, instead of using Shapley values as proposed by \citet{wei2024enhancing}\footnote{For $x_i$, if the unimodal predictions are correct with probabilities of 0.9 and 0.6, the Shapley values for both modalities would be the same, but the representational abilities are different.}, to avoid dataset expansion. 
For each sample $x_i$, we obtain unimodal prediction probabilities $p_i^k$, $k \in \{a, v\}$. The representational ability of each modality is then assessed by calculating the KL divergence between $p_i^k$ and a uniform distribution as 
\begin{equation}
\label{equ:KL}
    D_{KL}(p^k_i \| U) = \sum_{j=1}^{M} p^k_{i,j} \log \left( \frac{p^k_{i,j}}{u_j} \right),
\end{equation}

where $p^k_{i,j}$ represents the predicted probability for class $j$, and $u_j = \frac{1}{M}$ corresponds to the uniform distribution.
KL divergence measures the difference between two probability distributions. In the current application scenario, a smaller KL divergence indicates that the output is closer to a uniform distribution, meaning that the separability of the corresponding unimodal output is worse. This can be interpreted as insufficient training for the current modality on this sample.
Therefore, we decouple the multimodal inputs for each sample and only retain the modality that performs the worst, masking input from other modalities to achieve accurate unimodal training \cite{zhang2024multimodal,wei2024enhancing}. That is, the new training set is as follows:
\begin{equation}
\label{equ:newset}
    \mathcal{D'} = \left\{ x^{m_*} \mid x \in \mathcal{D}, m_* = \arg \min_{m_* \in K} D_{KL}(m_i) \right\}.
\end{equation}

By decoupling the multimodal inputs, the strong modality is exposed to fewer samples, while the weak modality is exposed to more, as shown in Figure \ref{fig:count}. This forces the model to focus more on learning the weak modality during the optimization process and prevents the strong modality from suppressing the weak modality's learning, thereby promoting modality balance, as shown in Figure \ref{fig:pho}, and ensuring more comprehensive multimodal learning.
More importantly, our sample-level unimodal evaluation method provides the basis for reassembling unimodal inputs to avoid modality clash.

\textbf{Reassemble Unimodal Data.}
After decoupling the multimodal inputs, we address modality laziness but find that modality clash remains, as shown in Figure \ref{fig:angle}. For more sufficient multimodal learning, we further analyze the phenomenon and identify that the interference is introduced at the batch level.

Without loss of generality, we assume that the multimodal fusion method selected for the model is Concatenation, as done by others  \cite{fan2023pmr, peng2022balanced, wei2024enhancing}. Let $\boldsymbol{W} \in \mathbb{R}^{M \times (d_{z^a} + d_{z^v})}$ and $\boldsymbol{b} \in \mathbb{R}^M$ represent the parameters of the linear classifier that produces the logits output. We can then express the output of the multimodal model (without softmax) as 
\begin{equation}
\label{equ:hx}
    f(\boldsymbol{x}) = \boldsymbol{W} \left [\phi^a(\theta^a, \boldsymbol{x}^{a}) ; \phi^v(\theta^v, \boldsymbol{x}^{v}) \right ] + \boldsymbol{b}.
\end{equation}

To analyze the optimization process of each modality, we represent $ \boldsymbol{W} $ as a block matrix composed of two parts: $ [\boldsymbol{W}^a, \boldsymbol{W}^v] $. We can then rewrite Equation \ref{equ:hx} as
\begin{equation}
\label{equ:hx2}
\begin{aligned}
f(\boldsymbol{x}) & = \boldsymbol{W}^a \left [\phi^a(\theta^a, \boldsymbol{x}^{a})\right] + \boldsymbol{W}^v \left [ \phi^v(\theta^v, \boldsymbol{x}^{v}) \right ] + \boldsymbol{b} \\
& = \boldsymbol{W}^a \boldsymbol{z}^a + \boldsymbol{W}^v \boldsymbol{z}^v + \boldsymbol{b}.
\end{aligned}
\end{equation}

Assuming that no additional classification heads are added for each modality branch, the loss for each batch is
\begin{equation}
\label{equ:lossle}
\begin{aligned}
    \mathcal{L}_{CE} & = - \frac{1}{\lvert B \rvert} \sum_{i=1}^{\lvert B \rvert} \log \left( \frac{e^{f(x_i)_{y_i}}}{\sum_{j=1}^{M} e^{f(x_i)_j}} \right) \\
    & = - \frac{1}{\lvert B \rvert} \sum_{i=1}^{\lvert B \rvert} \log \left( \frac{e^{{\boldsymbol{W}^a z_i^a + \boldsymbol{W}^v z_i^v + \boldsymbol{b}}_{y_i}}}{\sum_{j=1}^{M} e^{f(x_i)_j}} \right).
\end{aligned}
\end{equation}

Observing Equation \ref{equ:lossle}, we see that when using batch gradient descent, even if we decouple the multimodal inputs and mask the dominant modality for each sample, multimodal inputs still coexist within a batch. The inconsistency in optimization directions causes interference between their gradient update directions, ultimately leading to insufficient model training. Therefore, we propose that \textbf{cross-modal optimization interference originates at the batch level} and can be addressed by controlling the composition of each batch, referring to this as reassembling unimodal inputs.

We assume that the dataset $\mathcal{D}$ with $K$ modalities can be transformed into $\mathcal{D'}$ through decoupling multimodal inputs. Based on the retained modality, the dataset can be divided into $K$ subsets, denoted as
\begin{equation}
\label{equ:sub}
\mathcal{D}^{m_k} = \left\{ x_i \mid x_i^j=0, \forall j \neq k \right\}. 
\end{equation}

The divided subsets satisfy Equation \ref{equ:dataset}, ensuring that our training set does not expand.
\begin{equation}
\label{equ:dataset}
\begin{aligned}
& \bigcup_{k=1}^{K} \mathcal{D}^{m_k} = \mathcal{D'},  \\ 
&  \mathcal{D}^{m_i} \cap \mathcal{D}^{m_j} = \emptyset \quad \forall i \neq j
\end{aligned}
\end{equation}

To control the composition of each batch, we reassemble the unimodal inputs based on the decouple assignments and ensure that each batch $B_i $ contains data from only one subset, as shown in Equation \ref{equ:batch}.
\begin{equation}
\label{equ:batch}
B_i \subseteq \mathcal{D}^{m_k}, \quad k \in \{1, 2, \dots, K\}
\end{equation}

When sampling from a specific data subset, Equation \ref{equ:lossle} can be further simplified. For example, when sampling from the data subset corresponding to the video modality, Equation \ref{equ:lossle} simplifies to
\begin{equation}
\label{equ:lossle2}
    \mathcal{L}_{CE} = - \frac{1}{\lvert B \rvert} \sum_{i=1}^{\lvert B \rvert} \log \left( \frac{e^{{\boldsymbol{W}^v z_i^v + \boldsymbol{b}}_{y_i}}}{\sum_{j=1}^{M} e^{f(x_i)_j}} \right),
\end{equation}

where the phenomenon of cross-modal optimization interference has been effectively addressed.

In general, the pseudo-code for our method is provided in Algorithm \ref{alg:batch}. We bridge modality decoupling and reassembling through unimodal evaluation and implement the Data Remixing method to simultaneously tackle modality imbalance and insufficiency.
\begin{algorithm}[htb]
   \caption{Method of Data Remixing}
   \label{alg:batch}
\begin{algorithmic}
   \STATE {\bfseries Input:} input data $\mathcal{D} = \{(x^1_i, x^2_i,...,x^K_i), y_i\}_{i=1,2,...,N}$, number of modalities $K$, model parameters $\theta$, training epoch $E$ , warm-up epoch $E_r$
   \FOR{$e=0,\cdots, E-1$}
   \IF{$e < E_r$}
   \STATE Update model parameters $\theta$ with dataset $D$;
   \ELSE
   \FOR {$k=1,\cdots, K$}
   \STATE Initialize $\mathcal{D}^{m_k}:\mathcal{D}^{m_k} = \emptyset $;
   \ENDFOR
   \FOR{each sample $x$ in $D$}
   \STATE Calculate Uni-modal Ability $\{\varphi^1, \varphi^2, ..., \varphi^K\}$ with Equation \ref{equ:KL};
   \STATE Identify $k$ corresponding to the minimum $\varphi^k$;
   \STATE Mask $x^j, j \neq k$ to zero;
   \STATE Add $x$ into $\mathcal{D}^{m_k}$;
   \ENDFOR
   \FOR {$k=1,\cdots, K$}
   \STATE Update model parameters $\theta$ with dataset $\mathcal{D}^{m_k}$;
   \ENDFOR
   \ENDIF
   \ENDFOR

\end{algorithmic}
\end{algorithm}

\section{Experiments}\label{sec:experiments}

\subsection{Dataset and Experimental settings}
\textbf{CREMA-D} \cite{cremad} is an audiovisual dataset for emotion recognition, consisting of 7,442 video clips from 91 actors. The dataset includes six of the most common emotions: \textit{anger}, \textit{disgust}, \textit{fear}, \textit{happy}, \textit{neutral} and \textit{sad}. A total of 2,443 participants rated each clip for emotion and emotional intensity using three modalities: audiovisual, video only, and audio only. The entire dataset is randomly divided into a training and validation set of 6,698 samples and a test set of 744 samples, with a ratio of approximately 9:1.

\textbf{Kinetic-Sounds} \cite{ks} is a dataset derived from the Kinetics dataset \cite{kinetic}, which includes 400 action classes based on YouTube videos. Kinetic-Sounds consists of 31 action categories, selected for their potential to be represented both visually and aurally, such as playing various instruments. Each video is manually annotated for human actions using Mechanical Turk and is cropped to 10 seconds, focusing on the action itself. The dataset comprises 19k 10-second video clips, split into 15k for training, 1.9k for validation, and 1.9k for testing.

\textbf{Experimental settings.}
Unless otherwise specified, all feature extraction networks used in the experiments are ResNet-18 \cite{resnet}, trained from scratch. During training, we use the Adam \cite{kingma2014adam} optimizer with $ \beta = (0.9, 0.999) $ and set the learning rate to $ 5 \times 10^{-5} $. We obtain unimodal prediction results from the unimodal classification heads and also present the results of our method by masking specific modalities to zero, as in \citet{dropout} and \citet{wei2024enhancing}. All reported results are averages from three random seeds, with all models trained on two NVIDIA RTX 3090 GPUs using a batch size of 64.

\begin{table}[htbp]
\caption{Combination and comparison with conventional fusion methods. ``+ Remix'' indicates that our Data Remixing method is applied. \textbf{Bold} indicates that our method brings improvement.}
\label{tbl:compare1}
\vskip 0.15in
\begin{center}
\begin{small}
\begin{tabular}{p{2.8cm}|c|c}
\toprule
\textbf{Method} & \textbf{CREMAD} & \textbf{Kinetic-Sounds} \\
\midrule
Concatenation & 64.52\% & 50.23\% \\
Summation & 63.44\% & 51.66\% \\ 
Decision fusion & 67.47\% & 52.47\% \\
FiLM & 62.77\% & 49.61\% \\ 
Bi-Gated & 63.04\% & 49.92\% \\ 
\midrule
Concat + Remix & \textbf{72.72\%} & \textbf{55.63\%} \\
Sum + Remix & \textbf{71.51\%} & \textbf{54.09\%} \\
Decision + Remix & \textbf{70.70\%} & \textbf{54.86\%} \\
\bottomrule
\end{tabular}
\end{small}
\end{center}
\vskip -0.1in
\end{table}

\subsection{Comparison with conventional fusion methods}
We first compare our method with several representative multimodal fusion methods commonly used in deep learning frameworks: Concatenation (Concat) \cite{concat}, Summation (Sum), Decision Fusion (Decision) \cite{dicisionfuse}, FiLM \cite{film}, and Bi-Gated \cite{gated}. 
The results are shown in Table \ref{tbl:compare1}. We apply Data Remixing in combination with Concatenation as the representative fusion method for our approach. It is evident that our method significantly improves model performance across different datasets, with an increase of 8.20\% on CREMAD and 5.40\% on Kinetic-Sounds, outperforming other conventional fusion strategies. To further demonstrate the generalizability of our method, we combine Data Remixing with Summation and Decision Fusion, achieving substantial improvements across both datasets.

\subsection{Comparison with imbalanced multimodal learning methods}
Our method is primarily designed to address the issues of modality imbalance and insufficiency in multimodal learning. To evaluate the improvements, we compare our approach with several representative methods that target multimodal balance and sufficiency: G-Blend \cite{wang2020makes}, OGM-GE \cite{peng2022balanced}, Greedy \cite{greedy}, PMR \cite{fan2023pmr}, MLA \cite{mla} and Resample \cite{wei2024enhancing}. For a fair comparison, we adopt Concatenation as the fusion strategy for the baseline, following the approach used in the above methods. The results, including accuracy (Acc) and dataset expansion factor (Factor), are presented for all methods.

\begin{table}[htbp]
\caption{Comparison with other imbalanced multimodal learning methods. All modulation strategies are applied to the baseline, using Concatenation as the fusion method. We also include results of applying Data Remixing to Resample and MLA for further comparison.}
\label{tbl:compare2}
\vskip 0.15in
\begin{center}
\begin{small}
\begin{tabular}{p{2.6cm}|cc|cc}
\toprule
\multirow{2}{*}{\textbf{Method}} & \multicolumn{2}{c|}{\textbf{CREMAD}} & \multicolumn{2}{c}{\textbf{Kinetic-Sounds}} \\ 
& Acc(\%)& Factor & Acc(\%) & Factor\\
\midrule
Concatenation & 64.52 & 1 & 50.23 & 1 \\ 
\midrule
+ OGM-GE & 68.15 & 1 & 52.66  & 1\\
+ G-Blend & 69.89 & 1 & 53.55  & 1\\ 
+ Greedy & 68.28 & 1 & 51.20 & 1\\
+ PMR & 68.95 & 1 &  51.93 & 1 \\ 
+ MLA & 68.01 & 2 & 54.66 & 2\\ 
+ Resample  & 67.61 & 1.85 & 55.17 & 1.97 \\
\midrule
+ Remix (Ours)  & \textbf{72.72} & \textbf{1} & \textbf{55.63} & \textbf{1} \\
+ MLA + Remix & \textbf{74.19} & 1 & \textbf{57.75} & 1 \\ 
+ Resample + Remix & \textbf{73.25} & 1.92 & \textbf{58.40} & 2.01 \\ 
\bottomrule
\end{tabular}
\end{small}
\end{center}
\vskip -0.1in
\end{table}

Based on the results in Table \ref{tbl:compare2}, we observe that, due to the differing optimization trajectories of modalities, these imbalanced multimodal learning methods show performance improvements over traditional fusion methods. However, our method addresses both modality laziness and modality clash simultaneously, leading to even greater improvements.

Among these, the MLA and Resample strategies also perform relatively well, but hinder training efficiency. 
Under the same training conditions (especially GPUs and batch size), MLA employs a multi-stage training strategy and requires gradient modification based on previous features, making its training approximately \textbf{3$\times$ slower} than our method\footnote{Although MLA does not increase the number of samples, decoupling the inputs without selecting for unimodal training is similar to expanding the dataset, which slows down training.}.
The Resample method enhances the model's training on weak modality samples through resampling, which expands the training set by nearly twice, but it hinders training efficiency.
To more intuitively reflect the differences in efficiency, we measure the training time of four methods under the same conditions, as shown in Table \ref{tbl:sufficient}. We observe that balancing methods using unimodal single input tend to increase the training time, whereas Remix does not expand the training set, making it more efficient.
Unlike these methods, our approach neither expands the dataset nor reduces the model's training efficiency.

\begin{table}[htbp]
\caption{Training time to convergence (seconds) on CREMAD and Kinetic-Sounds datasets.}
\label{tbl:sufficient}
\vskip 0.15in
\begin{center}
\begin{small}
\begin{tabular}{p{2cm}|c|c|c|c}
\toprule
\textbf{Method} & \textbf{Baseline} & \textbf{Remix} & \textbf{Resample} & \textbf{MLA}\\
\midrule
CREMAD & 1536 & 2537 & 4525 & 6128 \\
Kinetic-Sounds & 3849 & 4946 & 10362 & 12868 \\
\bottomrule
\end{tabular}
\end{small}
\end{center}
\vskip -0.1in
\end{table}

Considering the similarities between our approach and the Resample and MLA methods—such as MLA’s decoupling of unimodal training, which indirectly achieves batch control, and Resample’s selective data training—we further combine Data Remixing with these methods to demonstrate its advancements and generalizability. We integrate a sample-level evaluation process into MLA, promoting model balance without increasing the number of input samples (which is why the Factor in Table \ref{tbl:compare2} is updated to 1), and reassemble the unimodal inputs after applying Resample. With our method applied, both approaches show significant improvements in their final results.

\subsection{Combination with complex cross-modal architectures}
The above methods and experiments are conducted using simple fusion methods, where multimodal fusion occurs after the unimodal encoders or classifiers. To validate the applicability of the Data Remixing method in more complex multimodal scenarios, we combine it with two intermediate fusion methods: MMTM \cite{joze2020mmtm} and CentralNet \cite{vielzeuf2018centralnet}. MMTM recalibrates the channel features of different CNN streams through squeezing and multimodal excitation steps, while CentralNet uses unimodal hidden representations alongside a central joint representation at each layer, performing fusion through a weighted summation learned during training.

As shown in Table \ref{tbl:compare3}, when multimodal fusion occurs during the encoding process, applying the Data Remixing method leads to significant improvements, further demonstrating the applicability of our method in complex cross-modal architectures.

\subsection{Analysis of Methods}
In this section, we provide a detailed analysis of the performance improvements introduced by our method. This includes comprehensive ablation experiments and an evaluation of the effectiveness of each design choice. Additionally, we explore different unimodal prediction methods to further demonstrate the broad applicability of the Data Remixing approach.

\begin{table}[tbp]
\caption{Results on CREMAD and Kinetic-Sounds with two types of complex cross-modal architectures. ``+ Remix'' indicates the application of our Data Remixing method.}
\label{tbl:compare3}
\vskip 0.15in
\begin{center}
\begin{small}
\begin{tabular}{p{2.8cm}|c|c}
\toprule
\textbf{Method} & \textbf{CREMAD} & \textbf{Kinetic-Sounds} \\
\midrule
Concatenation & 64.52\% & 50.23\% \\
Concat + Remix & \textbf{72.72\%} & \textbf{55.63\%} \\
\midrule
MMTM & 66.40\% & 52.27\% \\
MMTM + Remix & \textbf{68.82\%} & \textbf{54.78\%} \\
\midrule
CentralNet & 65.46\% & 54.09\% \\
CentralNet + Remix & \textbf{67.61\%} & \textbf{55.94\%} \\
\bottomrule
\end{tabular}
\end{small}
\end{center}
\vskip -0.1in
\end{table}

\subsubsection{Ablation Study}\label{sec:abla}
We conduct ablation studies to demonstrate the effectiveness of our two main designs, with the specific results shown in Table \ref{tbl:ablation}. Our experiments consist of two parts: (1) \textit{Decouple}, which involves decoupling the multimodal data and masking a specific modality without reassembling; (2) \textit{Reassemble}, which involves reassembling the modality inputs based on sample-level evaluation without specific training (i.e., without modality masking).

Observing the experimental results, we demonstrate the effectiveness of our method. In Experiment (1), we decouple the multimodal data and select specific samples for training. This approach facilitates modality balance, yielding some performance improvements. In Experiment (2), we reassemble the inputs based on sample-level evaluation, controlling the data composition within each batch. This strategy alleviates modality clash and leads to corresponding improvements. Finally, when both strategies are combined, the model's performance is further enhanced.

\begin{table}[htbp]
\caption{Results of ablation studies on CREMAD and Kinetic-Sounds. The second and third rows correspond to Experiment (1) and Experiment (2), respectively.}
\label{tbl:ablation}
\vskip 0.15in
\begin{center}
\begin{small}
\begin{tabular}{cc|c|c}
\toprule
\textbf{Decouple} & \textbf{Reassemble} & \textbf{CREMAD} & \textbf{Kinetic-Sounds} \\
\midrule
\XSolidBrush & \XSolidBrush & 64.52\% & 50.23\% \\
\Checkmark & \XSolidBrush & 69.89\% & 52.31\%\\ 
\XSolidBrush & \Checkmark & 68.68\% & 51.70\%\\ 
\Checkmark & \Checkmark & \textbf{72.72\%} & \textbf{55.63\%} \\ 
\bottomrule
\end{tabular}
\end{small}
\end{center}
\vskip -0.1in
\end{table}

\begin{figure*}[t]
\centering
    \subfigure[]{\includegraphics[width=0.33\textwidth]{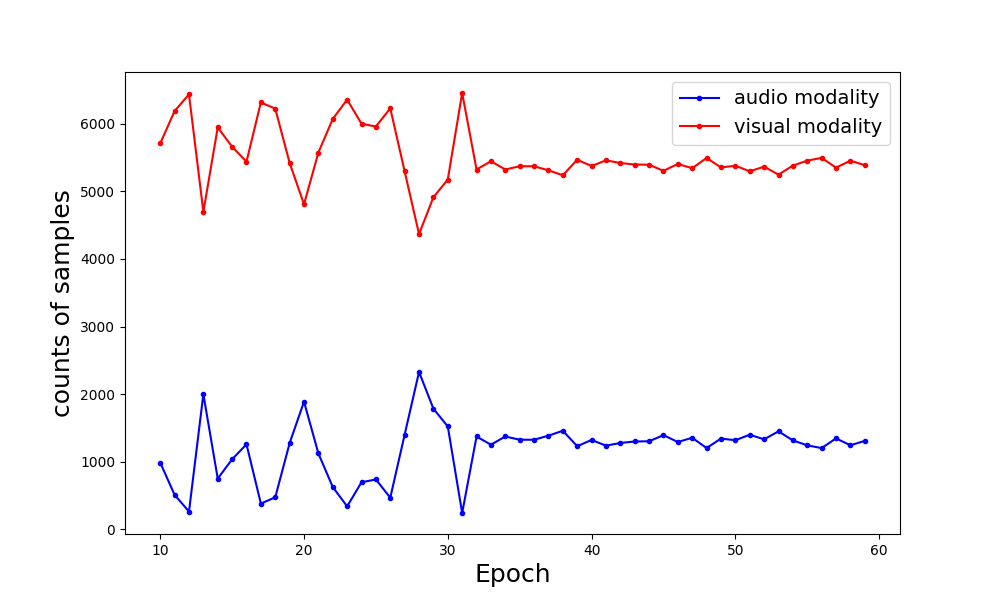}\label{fig:count}}
    \subfigure[]{\includegraphics[width=0.33\textwidth]{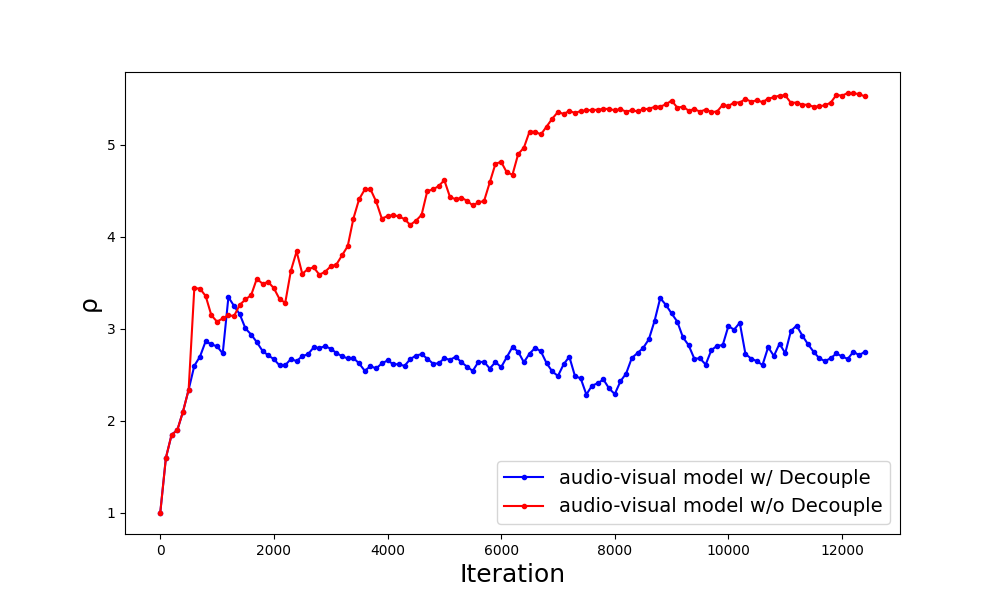}\label{fig:pho}}
    \subfigure[]{\includegraphics[width=0.33\textwidth]{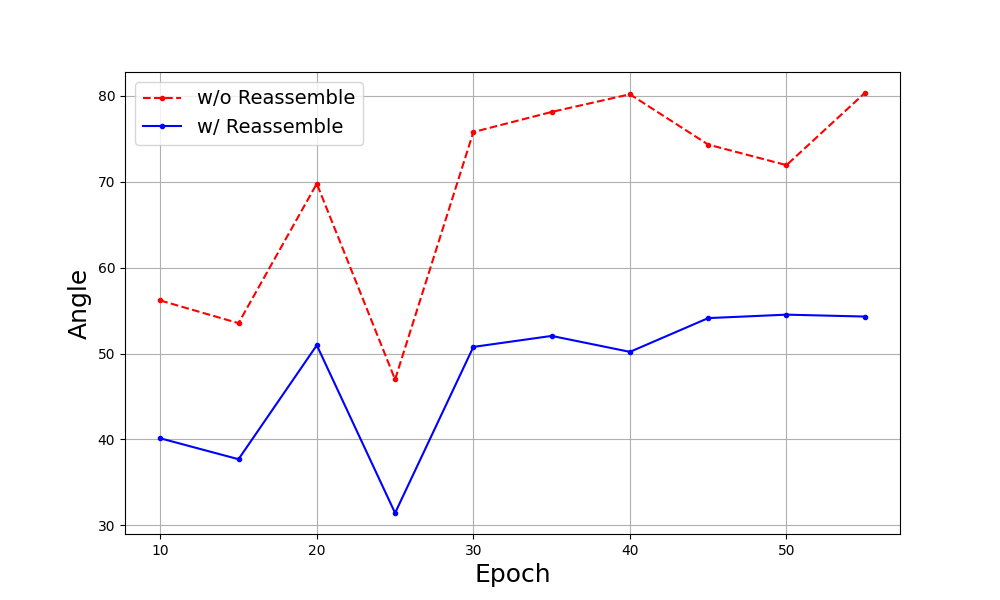}\label{fig:angle}}
\caption{Proof of effectiveness of Decouple and Reassemble Methods. The results are obtained on CREMAD. (a) Statistics of the number of samples used for training specific modalities. (b) The change of the imbalance ratio $\rho$. (c) Comparison of gradient direction discrepancies.}
\label{fig:umap}
\vskip -0.2in
\end{figure*}

\subsubsection{Study of Decouple Multimodal Data}
In the process of conducting sample-level evaluation and decoupling multimodal data, we choose to mask the better-performing modality and train with specific samples to alleviate modality imbalance.
Figure \ref{fig:count} shows the distribution of samples assigned to different modalities during training. It is clear that the strong modality consistently receives fewer samples compared to the weak modality. Figure \ref{fig:pho} illustrates the change in imbalance ratio $\rho$ \cite{peng2022balanced} before and after decoupling the modality inputs. The results indicate that our method alleviates the modality imbalance, supporting our hypothesis that balancing modality representation via sample quantity and difficulty can improve model performance.
Note that our method is applied after a 10-epoch warm-up stage.

\subsubsection{Study of Reassemble Unimodal Data}
To mitigate cross-modal optimization interference, we reassemble unimodal inputs to control the data composition within each batch. Our theoretical analysis of batch control supports the effectiveness of this strategy, and the improvements in unimodal performance are evident. Specifically, audio accuracy on CREMAD increased by 1.75\%, while video accuracy improved by 2.96\%, since the strong modality is less affected compared to the weak modality. 

To further validate the effectiveness of our method in alleviating gradient interference, we evaluate the gradient update discrepancies with and without data reassembling, as shown in Figure \ref{fig:angle}. Specifically, we assess the gradient update discrepancies of the audio modality (the strong modality) on CREMAD at different training stages. We calculate the angle between the actual gradient update direction and the ideal guidance direction for the audio modality. The results show that after reassembling the unimodal inputs, the gradient update direction aligns more closely with the ideal direction, providing further evidence of the effectiveness of our approach.

\begin{table}[htbp]
\caption{Results of accuracy(\%) on CREMAD and Kinetic-Sounds using different methods for unimodal predictions.}
\label{tbl:unimodal}
\vskip 0.15in
\begin{center}
\begin{small}
\begin{tabular}{p{2.3cm}|cc|cc}
\toprule
\multirow{2}{*}{\textbf{Method}} & \multicolumn{2}{c|}{\textbf{CREMAD}} & \multicolumn{2}{c}{\textbf{Kinetic-Sounds}} \\
& Dropout & Head & Dropout & Head \\
\midrule
Concatenation & 61.56 & 64.52 & 49.58 & 50.23 \\
Summation  & 59.68 & 63.44 & 48.00 & 51.66 \\
Decision fusion & 62.23 & 67.47 & 50.08 & 52.47   \\
\midrule
Concat + Remix & \textbf{70.03} & \textbf{72.72} & \textbf{53.89} & \textbf{55.63}   \\
Sum + Remix & \textbf{68.68} & \textbf{71.51} & \textbf{53.12} & \textbf{54.09}   \\
Decision + Remix & \textbf{68.55} & \textbf{70.70} & \textbf{53.55} & \textbf{54.86}   \\
\bottomrule
\end{tabular}
\end{small}
\end{center}
\vskip -0.1in
\end{table}

\subsubsection{Study of Unimodal Prediction Methods}\label{sec:unimodal}
Since our method requires accurate unimodal predictions, we add a classification head to each modality branch and synchronize the updates of the classification head parameters by modifying the loss function. This approach has been shown to improve model performance \cite{wang2020makes}. To further validate our method, we compare the results with the dropout method \cite{dropout, wei2024enhancing} for unimodal prediction evaluation across three traditional fusion strategies. The results are presented in Table \ref{tbl:unimodal}. From the table, we observe that both methods of obtaining unimodal predictions show effective improvements when applying our strategy. Notably, regardless of the chosen method, our approach does not introduce any additional overhead during inference.

\section{Conclusion}
Multimodal models often face challenges due to the differences in optimization trajectories between modalities, leading to issues of insufficient and imbalanced learning during joint training. We propose the Data Remixing method, which decouples multimodal data based on unimodal separability and reassembles unimodal data to ensure consistency between modalities at the batch level.
Our method effectively mitigates both modality laziness and modality clash, achieving significant performance improvements across various datasets, fusion methods, and model structures. Additionally, it does not require dataset expansion or introduce extra computational overhead during inference.
Moreover, to the best of our knowledge, we are the first to propose that the challenges of multimodal learning originate at the batch level and to offer a solution for them.
However, a limitation of our approach arises when one modality serves primarily as an auxiliary modality with limited information \cite{wei2025diagnosing}. In such cases, the unimodal evaluation and allocation strategy may require further refinement.

\section*{Acknowledgments}
The work is jointly supported by the National Natural Science Foundation of China (NSFC) under Grant 62261160576 and Grant 62203024, the Beijing Natural Science Foundation (4252026), the Research and Development Program of Beijing Municipal Education Commission (KM202310005027), and the Fundamental Research Funds for the Central Universities of China. This research work is also supported by the Big Data Computing Center of Southeast University.

\section*{Impact Statement}
This paper presents work whose goal is to advance the field of Machine Learning. There are many potential societal consequences of our work, none of which we feel must be specifically highlighted here.

\nocite{yang2022mcl,umap,ghorbani2020neuron,van2008visualizing}

\bibliography{main}
\bibliographystyle{icml2025}

\end{document}